\documentclass[10pt,twocolumn,letterpaper]{article}

\usepackage[pagenumbers]{cvpr}      %

\usepackage{cuted} %

\usepackage{currfile} %

\usepackage{caption} %

\usepackage{microtype}

\usepackage{pifont}

\usepackage{booktabs}
\usepackage{multirow}
\usepackage{graphicx}
\usepackage{amsmath,amssymb}
\usepackage{tikz}
\usetikzlibrary{arrows.meta,positioning,fit,calc}
\usepackage{xspace}
\newcommand{\SSv}{ScreenSpot-v2\xspace}
\newcommand{\MW}{Mind2Web\xspace}

\definecolor{cvprblue}{rgb}{0.21,0.49,0.74}
\usepackage[pagebackref,breaklinks,colorlinks,allcolors=cvprblue]{hyperref}

\title{Right Click, Wrong Box: Grading GUI Grounding by What a Click Does}

\newcommand{\PaperAuthorNames}{Xiaokang Yuan; Binhui Yao; Luwei Yang; Ruimin Chen; Zeng Wang}
\author{%
  Xiaokang Yuan\textsuperscript{1}\quad
  Binhui Yao\textsuperscript{1}\quad
  Luwei Yang\textsuperscript{2}\quad
  Ruimin Chen\textsuperscript{2}\quad
  Zeng Wang\textsuperscript{2}\\[5pt]
  {\small \textsuperscript{1}Zhejiang University, Hangzhou, China}\\
  {\small \textsuperscript{2}Zhejiang Geely Holding Group Co., Ltd, Hangzhou, China}\\[3pt]
  {\small \href{mailto:22621230@zju.edu.cn}{\texttt{22621230@zju.edu.cn}}}%
}
\date{}

\hypersetup{pdftitle={Right Click, Wrong Box: Grading GUI Grounding by What a Click Does},pdfauthor={\PaperAuthorNames}}
\begin{document}
\twocolumn[{%
\renewcommand\twocolumn[1][]{#1}%
\maketitle
\begin{center}
  \centering
  \captionsetup{type=figure}
  \includegraphics[width=\textwidth]{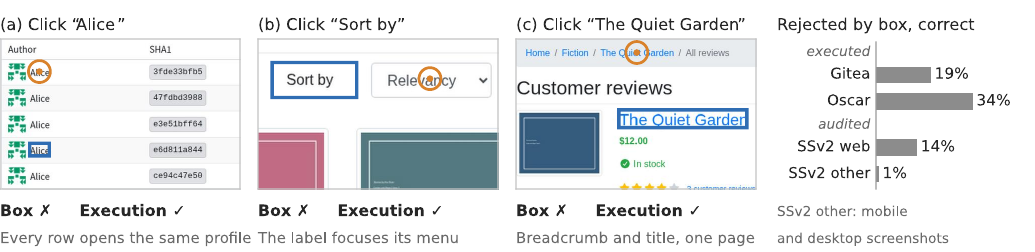}
  \vspace{-1.2em}
  \captionof{figure}{\textbf{Correct clicks that box scoring rejects.} (a--c) Predictions of GTA1-7B (a, b) and
  UI-Venus-1.5-8B (c) on Gitea and the Oscar shop. The blue box is the annotation and the orange circle the model's
  click. Each click misses the box, and executing it produces the same outcome as clicking the target. Instructions
  are shortened from the template \texttt{Click the <role> "<text>"}. Right: share of box-rejected clicks that reach
  the target outcome, by execution for three 7--8B grounders and by a reviewed audit of 11 models on \SSv.}
  \label{fig:teaser}
\end{center}%
\vspace{0.6em}
}]
\begin{abstract}
Leading GUI grounders score 93 to 96\% on \SSv, and reported gaps between new and prior models are about a point. At this level
the useful question is what the remaining errors are. Benchmarks count a click as correct only inside one annotated
box, but interfaces often expose one action through several elements: a duplicated link, the label of a checkbox, the
title next to a logo. The box is not wrong. The rule is incomplete. We call a click correct when executing it produces
the same outcome as clicking the annotated target. A probe that executes and fingerprints about a thousand clicks per
page turns this definition into outcome maps and an execution judge for live web applications, with no new
annotation. We execute the predictions of three strong 7--8B grounders on a code forge and an online shop. About one in five
and one in three of the clicks that box scoring rejects reach the intended outcome, and the shares remain one in seven
and one in three when the instructions are rewritten as natural tasks. A blind audit of 11 models on \SSv (three
7--8B grounders and eight 3B variants) finds the same effect on web screenshots, with two independent reviewers on a
60-unit subset. The share is highest for the strongest grounders. Training seeds and test-set sampling add noise
that more runs and paired intervals control. Box scoring instead under-counts correct clicks, which repetition does
not remove, and the audit-based correction of 0.2 to 1.8 points on \SSv is as large as the gaps the field reports.
The correction is needed in evaluation. In our 3B GRPO study, crediting equivalent clicks in the reward gives no
stable gain over box rewards across three seeds and four test sets. We release the probe, the outcome maps, the audit and
every prediction.
\end{abstract}

\section{Introduction}
\label{sec:intro}

GUI agents act through clicks, so grounding an instruction to a screen location is the perception step every agent
depends on. The strongest open grounders we evaluate score 93 to 96\% on \SSv~\cite{wu2024osatlas}
(\cref{sec:eval}), and across 29 recent grounding papers the median absolute gap between a new model and the strongest prior model of
its size is 1.15 points. At this level the useful question is no longer how many clicks miss the box. It is what the remaining
misses are. When the remaining mistakes of top ImageNet models were reviewed one by one, nearly half turned out not
to be mistakes~\cite{vasudevan2022dough}.

Grounding benchmarks score a predicted point with one rule. It is correct when it falls inside the annotated box of the
target element~\cite{cheng2024seeclick,wu2024osatlas,li2025screenspotpro,xie2025jedi,deng2023mind2web}, and
reinforcement learning for grounding uses the same rule as its reward~\cite{yang2025gta1,tang2025guig2}. The rule
assumes that each instruction has one correct region. Interfaces are built the other way. A commit list links every author name to the same
profile, a form label focuses its control, and a shop shows a product's name in the breadcrumb and in the title, both
linking to the product (\cref{fig:teaser}). Public screenshots show the same pattern (\cref{fig:ssv2}). None of this is an annotation mistake. The box is not wrong. The rule is
incomplete, because a click outside the box can do exactly what the instruction asks. The converse also happens, and
parts of an annotated box can be inert.

We define correctness by what a click does. Given an interface state and an annotated target, a click is
\emph{outcome-correct} when executing it produces the same outcome as clicking the target. A probe executes a click in
a browser, intercepts every request that would change server state, and records a fingerprint of the result. Running
the probe densely over a page yields an \emph{outcome map}, the set of pixels that achieve each element's outcome.
Running it on model predictions yields an \emph{execution judge}. Neither needs annotation beyond the existing box. For
public screenshot benchmarks, where nothing can be executed, we run a blind audit with the same definition.

\textbf{What the remaining errors are.} On a self-hosted Gitea code forge and a Django Oscar shop, 17.7\% and 31.2\%
of live targets have pixels outside the box that reach the same outcome or pixels inside it that do nothing. Executing the predictions of UI-Venus-1.5-8B, GTA1-7B and Qwen3-VL-8B shows that about one in five clicks rejected by
box scoring is outcome-correct on Gitea, and about one in three on Oscar. Rewriting the instructions as natural
tasks keeps these shares at 14.3\% and 34.8\%. On \SSv, a blind audit of 11 models,
cross-checked by two independent reviewers, finds 6.4\% of rejected clicks outcome-correct overall and 14.4\% on web
screenshots. Among the audited models the share is highest for the strongest grounders, so box scoring misreads more of the errors that still separate strong grounders
than of the errors of weaker models.

\textbf{Bias, not noise.} Another training seed moves \SSv accuracy by up to 2.6 points, and test-set sampling limits
paired comparisons of 7--8B models to about 1.4 points. Both are noise that more seeds and a paired interval control,
yet none of the 29 papers reports run-to-run variation on \SSv. Box scoring adds a bias, a one-sided under-count of
the correct clicks of every model. It is estimated at 0.2 to 1.8 points on \SSv, and no number of runs or test items
removes it.
That is the size of the gaps the field reports, and the part that differs between models changes the point-estimate
order of models within a point of each other.

\textbf{Where the fix belongs.} In our 3B study, a GRPO reward that credits the equivalent clicks found by the probe
brings no stable gain over box rewards across three seeds and four test sets. Trained models already reach the target
outcome on every held-out item whose target has an exact duplicate, so the correction is needed in evaluation, while
its value as a training reward remains unshown at this scale.

Our contributions are:
\begin{itemize}
  \item A definition of outcome-defined correctness for GUI grounding, and a probe that turns it into outcome maps and
  an execution judge for live web applications without new annotation.
  \item An audit of what box scoring rejects on two live applications and two public benchmarks, with execution
  judging of strong grounders and a reviewed screenshot audit of 11 models.
  \item An error budget for one-point comparisons on \SSv that separates the scoring bias from seed and sampling
  noise, with reporting recommendations.
  \item A controlled training study that places the fix in evaluation rather than in the reward. All maps, judgments,
  predictions and protocols are released.
\end{itemize}

\section{Related Work}
\label{sec:related}

\textbf{GUI grounding benchmarks and models.} ScreenSpot~\cite{cheng2024seeclick} established the screenshot grounding
task across mobile, desktop and web, and \SSv~\cite{wu2024osatlas} re-annotated 11.3\% of its samples to fix wrong
boxes and ambiguous instructions. ScreenSpot-Pro~\cite{li2025screenspotpro} raises resolution and domain difficulty,
OSWorld-G~\cite{xie2025jedi} adds fine-grained annotations, and \MW~\cite{deng2023mind2web} provides real web pages
with element-level ground truth. All of them score a prediction by its position relative to one annotated region.
Grounding models have grown from supervised pretraining on large GUI corpora~\cite{cheng2024seeclick,wu2024osatlas,gou2025uground,lin2025showui,xu2025aguvis,qin2025uitars}
to coordinate-free heads~\cite{wu2025guiactor}, data-centric recipes~\cite{zhang2025phiground} and reinforcement
learning, as in GTA1~\cite{yang2025gta1}, GUI-G$^2$~\cite{tang2025guig2} and UI-Venus~\cite{venus2026uivenus15}.
They and general models such as Qwen-VL~\cite{bai2025qwen25vl,bai2025qwen3vl} report results under this protocol.

\textbf{Rewards for grounding.} GRPO~\cite{shao2024deepseekmath} with a point-in-box reward is the standard recipe for
RL-trained grounders~\cite{yang2025gta1}, and GUI-G$^2$ replaces the binary hit with a Gaussian reward around the box
centre~\cite{tang2025guig2}. GUI-Libra~\cite{yang2026guilibra} observes that offline GUI rewards are only partially
verifiable, since a zero reward may hide an alternative valid action, and proposes KL regularisation and success-adaptive
scaling of negative advantages, studied mainly for multi-step navigation. We measure how often such alternative
actions exist in single-step grounding, give the reward access to those the probe finds, and train GUI-Libra alongside.

\textbf{Multiple valid actions.} MobiBench~\cite{im2025mobibench} annotates several valid actions per step for mobile
agents and shows that single-path evaluation underestimates task success. Concurrent work, ConClick~\cite{zhang2026conclick}, studies
grounding errors that land on non-target elements and uses a vision-language judge to estimate that at most 6\% of
them are valid alternatives. Its sample and categories differ from ours, so the comparison is only broad. Our \SSv
audit adds three things: human
judgments with measured agreement, a breakdown by platform that reveals a much larger share on the web, and execution
judging on two live applications where equivalence is observed directly.

\textbf{Execution-based evaluation.} AutoGUI~\cite{li2025autogui} executes interactions and asks an LLM to describe each
element's function from the change on screen, to create training data. We compare outcomes of clicks to grade
predictions. WebArena~\cite{zhou2024webarena} and OSWorld~\cite{xie2024osworld} judge agents
by the state they reach at the end of a task. We bring the same principle down to a single click, which makes it
usable for grounding benchmarks and grounding rewards.

\textbf{Auditing benchmarks.} Re-examining test sets has repeatedly changed how the vision community reads its
leaderboards, from ImageNet re-labelling and replication~\cite{beyer2020imagenet,recht2019imagenet} to systematic
label-error audits~\cite{northcutt2021pervasive}. A review of every remaining mistake of top ImageNet models found that
nearly half were not mistakes~\cite{vasudevan2022dough}, and Boundary IoU showed that the standard mask measure
overlooks improvements in boundary quality~\cite{cheng2021boundaryiou}. Work on benchmark
variance~\cite{bouthillier2021variance,henderson2018deeprl,agarwal2021precipice,dror2018hitchhiker} shows that seeds
and test-set sampling can rival the effects being claimed. In GUI grounding, the re-annotation behind \SSv corrected
individual labels. We keep the labels and audit the grading rule, with ground truth obtained by execution wherever the interface can
run, and we measure seed and sampling variance alongside it.

\section{Outcome-Defined Correctness}
\label{sec:method}

\subsection{Definition}
Let $s$ be an interface state, $p$ a screen point and $o(s,p)$ the outcome of clicking $p$ in state $s$. For an
annotated target with box $b$, the target outcome $o_b$ is the most frequent non-null outcome over points inside $b$,
with the box centre breaking ties. A click is outcome-correct when $o(s,p) = o_b$, and the outcome-equivalent region is
$E(s,b) = \{p : o(s,p) = o_b\}$. Box scoring replaces $E(s,b)$ with $b$. The two disagree in two ways. Points in
$E(s,b) \setminus b$ are correct clicks that box scoring rejects, and points in $b \setminus E(s,b)$ are inert clicks
that it accepts. Outcomes are defined relative to the state $s$, which is fixed by the screenshot or page being evaluated. On live
pages, $o(s,p)$ is the operational fingerprint of \cref{sec:probe}.

\subsection{Executing and fingerprinting a click}
\label{sec:probe}
\begin{figure}[t]
  \centering
  \includegraphics[width=196.59033pt]{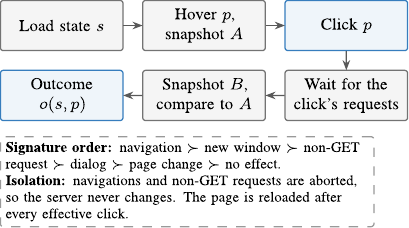}
  \caption{\textbf{The probe.} Each screen point $p$ is executed in the interface state $s$ and fingerprinted by
  the first matching signature. Running it over a grid gives an outcome map, and running it on a model's prediction
  gives an execution judgment.}
  \label{fig:probe}
\end{figure}

The probe (\cref{fig:probe}) runs Chromium at $1280\times720$. For each point it hovers for 120\,ms, records the page state, clicks, waits
until requests issued by the click have settled (at least 250\,ms and at most 1.5\,s beyond that, ignoring long-lived
streaming connections), and records the state again. The outcome is the first matching signature in a fixed order:
main-frame navigation (method, normalised URL and form fields), a new window, a non-GET request, a dialog, a change in
the visible page, or no effect. Navigations and non-GET requests are intercepted and aborted, so the server state
never changes and every click starts from the same page. Hover effects cancel out because both snapshots include them,
and nodes that change on their own, such as clocks, are found from idle snapshots and excluded. A text field gaining
focus counts as an outcome, and other elements gaining focus do not. After any click with an effect the page is
reloaded.

\subsection{Outcome maps}
\label{sec:maps}
For every visible interactive element the probe clicks a $3\times3$ grid inside the box and one point 3\,px outside
each edge, and it adds a 48\,px grid over the whole viewport (at most 1,500 points per page). Elements whose box
contains no effective point are \emph{dead} and set aside. For each live target we record whether a different element
reaches its outcome (a \emph{distinct equivalent}, such as a duplicated link or a label), whether at least 10\% of its
box is inert, and whether clicks just outside the box still reach its outcome. A target \emph{disagrees} with box
scoring if any of the three holds. A random 10\% of points on every page is re-probed from a fresh load to measure
repeatability.

We probe two self-hosted applications with different interaction structure, both seeded deterministically. Gitea
1.25 is a code forge similar to the GitLab site of WebArena~\cite{zhou2024webarena}, with users, organisations,
repositories, issues, pull requests and wikis (94 pages). Django Oscar 4.2 is an open-source e-commerce framework,
close to the shopping site of WebArena, with a storefront of product grids, categories, reviews, basket and account
pages and a staff dashboard of tables and forms (60 pages). On Oscar an application-side middleware rolls back every
database transaction, so that page views leave no trace. Both databases are identical before and after probing.

\subsection{Execution judging}
\label{sec:exec}
Each live target with visible text becomes a grounding item with the instruction template
\texttt{Click the <role> "<text>"}, up to 40 items per page. We execute each model's predicted point with the same probe
and compare its outcome with the target outcome. As a reference we also execute the centre of every annotated box. The
centre reproduces the mapped outcome on 98.1\% of items, which bounds environment drift between mapping and judging.
Confidence intervals resample pages~\cite{efron1993bootstrap}.

\subsection{Screenshot audit}
\label{sec:audit}
Public benchmarks ship screenshots, so outcomes cannot be executed. For \SSv we use human judgment with the same
definition. We take the existing predictions of 11 models and keep the parsed clicks that miss the box. Clicks on the same item
that lie within 0.8\% of the long image side are merged, giving 1,036 audit units. A fixed random order selects a
40-unit pilot and a 360-unit main sample. Each unit is shown as a panel with the full screenshot and a zoomed crop,
the target box in red and the click in blue, together with the instruction. Model names are hidden. A unit is
\emph{E-same} when the click hits the target control outside its box (the box covers only an icon of a larger button),
\emph{E-dup} when it hits another control with the same function (a duplicated link, the label of a checkbox or field),
\emph{N} when the function differs, and \emph{U} when the screenshot does not show enough to decide. Clear conventions
are required to accept a whole row or card as clickable, and uncertain cases go to U. Two independent reviewers judge a
random 60-unit subset of the main sample without seeing our labels.

All decision rules, thresholds and sampling seeds were recorded with timestamps and checksums before the data they
govern was produced. The supplementary material lists the rule files behind every analysis in this paper and each later
amendment, together with what had been seen when it was written.

\section{What Box Scoring Misreads}
\label{sec:eval}

Every instrument in this study is checked against an independent repetition or an independent judge.
\Cref{tab:checks} collects these checks, and all of them pass with a wide margin.

\begin{table}[t]
  \centering
  \small
  \setlength{\tabcolsep}{2.5pt}
  \begin{tabular}{@{}llr@{}}
    \toprule
    Instrument & Check & Result \\
    \midrule
    \multirow{4}{*}{Probe} & Same outcome after fresh reload & 97.9 / 99.8\% \\
    & Same outcome across probe versions & 99.4\% \\
    & Disagreements confirmed in self-check & 28/30 \\
    & Server database after probing & unchanged \\
    \midrule
    \multirow{2}{*}{Judge} & Box centre reaches mapped outcome & 98.1 / 98.7\% \\
    & Same verdict on re-execution & 98.6 / 98.7\% \\
    \midrule
    \multirow{2}{*}{Audit} & Reviewer A vs reviewer B, $\kappa$ & 0.95 \\
    & Ours vs reviewers, $\kappa$ & 0.89, 0.84 \\
    \bottomrule
  \end{tabular}
  \caption{\textbf{Checks on each instrument.} Pairs of values are Gitea / Oscar. Probe repeatability is the mean per page
  of a random 10\% of points re-probed from a fresh load. Version agreement covers 60,995 points on 58 pages. Re-execution covers 712
  predictions. Cohen's $\kappa$~\cite{cohen1960kappa} is computed over E, N and U on 60 units. Details are in the
  supplement.}
  \label{tab:checks}
\end{table}

\subsection{Outcome maps of live applications}
The probe executed about 1,000 clicks on each of 94 Gitea pages and about 700 on each Oscar page. Of 4,970 live targets, 17.7\% disagree with their
box. Most disagreements come from distinct equivalents (15.1\% of targets). Duplicated links form the largest group,
such as commit hashes and messages, issue titles and numbers, and user names and avatars. Labels that focus or toggle
their control come second (\cref{tab:types}). Smaller groups are elements that stay clickable beyond their box
(\cref{fig:outmap}c in the supplement) and boxes with at least 10\% inert area (2.6\% of targets). The estimate is stable. Restricting
the analysis to pages with at least 95\% repeatability leaves the rate at 17.4\%. Discarding every equivalence that
rests on a page change without a label, the type that contains one of the two self-check false positives, still
leaves 16.1\%. The shop shows the same pattern more strongly. On 56 Oscar pages that meet the repeatability
threshold, 31.2\% of 1,224 live targets disagree with their box, mostly through distinct equivalents (27.5\%), for example the image and the title of a product card that open the
same product page.

\begin{table}[t]
  \centering
  \small
  \begin{tabular}{@{}lr@{}}
    \toprule
    Type & Targets \\
    \midrule
    Duplicate link (navigation to the same URL) & 566 \\
    Label and its control & 86 \\
    Other page-change equivalent & 60 \\
    At least 10\% inert area in the box & 64 \\
    Clickable area beyond the box & 69 \\
    \midrule
    Any disagreement & 774 \\
    \bottomrule
  \end{tabular}
  \caption{Disagreeing targets by type on the 85 pages with at least 95\% repeatability (4,436 live targets). A target
  can have several types.}
  \label{tab:types}
\end{table}

\subsection{Execution-judged accuracy}
We judged 3,586 Gitea items and 1,147 Oscar items by execution for three strong grounders:
UI-Venus-1.5-8B~\cite{venus2026uivenus15}, GTA1-7B~\cite{yang2025gta1} and Qwen3-VL-8B~\cite{bai2025qwen3vl}.
\Cref{tab:exec} shows the result. About one in five clicks that box scoring rejects reaches the target outcome on
Gitea, and about one in three on Oscar. Accuracy rises by about two points for every model on both applications.
Box-accepted clicks with a different outcome occur at the same rate as for the box centre itself, so inert pixels
inside boxes play a minor role for these models.

\begin{table}[t]
  \centering
  \small
  \setlength{\tabcolsep}{3.5pt}
  \begin{tabular}{@{}lcccc@{}}
    \toprule
    Model & Box & Outcome & $\Delta$ [95\% CI] & Rescued \\
    \midrule
    \multicolumn{5}{@{}l}{\emph{Gitea (3,586 items)}} \\
    UI-Venus-1.5-8B & 82.3 & 84.3 & +2.0 [0.6, 3.3] & 130 (20.5\%) \\
    GTA1-7B         & 81.1 & 83.2 & +2.2 [0.9, 3.4] & 135 (19.9\%) \\
    Qwen3-VL-8B     & 79.7 & 81.9 & +2.2 [0.9, 3.5] & 128 (17.6\%) \\
    Box centre      & 100  & 98.1 &                  & 0 \\
    \midrule
    \multicolumn{5}{@{}l}{\emph{Oscar (1,147 items)}} \\
    UI-Venus-1.5-8B & 91.3 & 93.0 & +1.7 [0.6, 2.9] & 34 (34.0\%) \\
    GTA1-7B         & 90.1 & 92.6 & +2.4 [1.2, 3.6] & 43 (38.1\%) \\
    Qwen3-VL-8B     & 90.3 & 92.3 & +2.0 [1.0, 3.0] & 33 (29.7\%) \\
    Box centre      & 100  & 98.7 &                  & 0 \\
    \bottomrule
  \end{tabular}
  \caption{\textbf{Execution judging on two live applications.} Accuracy (\%) under box scoring and outcome scoring.
  ``Rescued'' counts box-rejected clicks that reach the target outcome, with their share of all box-rejected
  clicks. The box centre is a reference click; its agreement with the mapped outcome bounds environment drift. Intervals resample pages.}
  \label{tab:exec}
\end{table}

The gain concentrates where the interface offers several routes to the same action. On the 193 items whose
instruction matches several elements with the same outcome, models often pick the other copy, and box accuracy stays
at 28.5--35.2\%. Outcome accuracy on these items is 90.7--100\%. On the 3,092 items with a unique match, outcome
accuracy sits slightly below box accuracy, by less than the gap of the box centre itself. The 301 items whose matching
elements lead to different outcomes stay hard under both scores, so execution judging credits genuine equivalence
only. Oscar repeats this picture: on its 55 items with same-outcome copies, box accuracy is 47--49\% and outcome
accuracy 96--100\%, and on unique items the two scores differ by less than a point. On Oscar, GTA1-7B trails
Qwen3-VL-8B by 0.17 points under box scoring and leads by 0.26 under outcome scoring.

\textbf{Natural instructions.} The template names the element's text, so an element and its same-text copies match the
instruction equally well. To test whether the effect depends on this, a text-only model (Qwen3-4B-Instruct-2507)
rewrote every instruction as a task a user would give, from the element's text, type and effect and without the
screen. For example, \emph{Click the link ``1.25.0''} became \emph{Go to the admin configuration page}. In a reading
of 100 random rewrites, 10 were clearly unfit, mostly checkboxes described as text fields and invented search terms.
The task becomes harder, and box accuracy drops by 10 to 16 points. The share of box-rejected clicks that reach the
target outcome stays high, at 14.3\% on Gitea (19.2\% with the template) and 34.8\% on Oscar (34.0\%), pooled over
the three grounders. A rule written before any rewrite existed counted the effect as persisting if both shares stayed
above half their template values. Outcome scoring now raises accuracy by 2.5--3.2 points on Gitea and 6.2--8.4 points
on Oscar. Rescued clicks also appear on items whose template text was unique, 6.8\% of rejections on Gitea and
30.6\% on Oscar, so the effect is not limited to literal copies of the instruction text (\cref{tab:nl}).

\begin{figure*}[t]
  \centering
  \includegraphics[width=\textwidth]{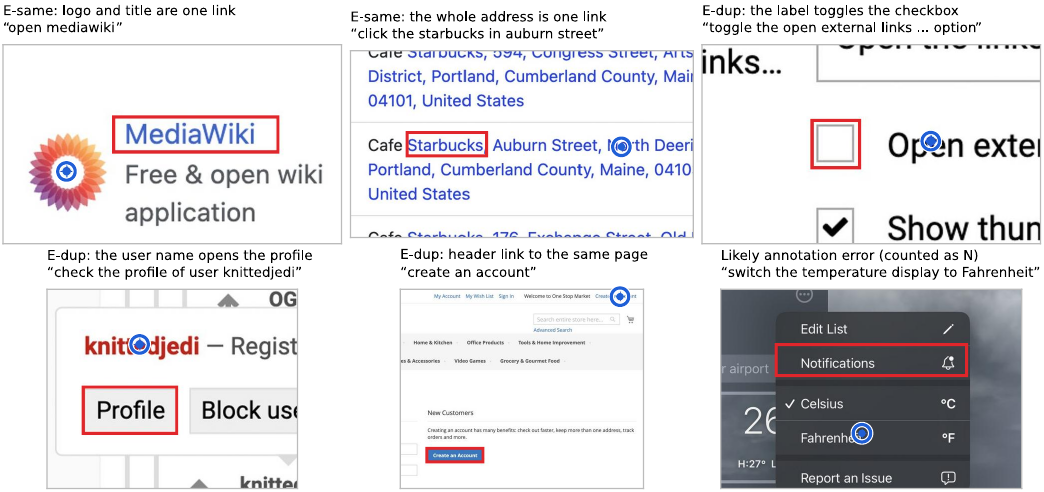}
  \caption{\textbf{Outcome-correct clicks that \SSv rejects.} Zoomed audit panels: red box is the annotation, blue
  circle the model click, instruction in quotes. E-same clicks hit the target control outside its box, E-dup clicks hit
  another control with the same function. In the last panel the click follows the instruction and the box marks a
  different control. We release such cases as likely annotation errors and count them as N.}
  \label{fig:ssv2}
\end{figure*}

\subsection{Public benchmarks}
\textbf{\MW.} The released \MW pages strip almost all element attributes, link targets included, so equivalence
can only be recovered from label relations. Such static equivalents are visible for 1.3\% of the 4,838 click steps we
evaluate. Box and outcome accuracy of the three grounders therefore differ by at most 0.26 points. Where an equivalent is visible the effect is
large. With task-level instructions, outcome accuracy on these steps exceeds box accuracy by 12.6 to 20.3 points.
Duplicated links, the largest group on Gitea, are invisible in such snapshots, which is why live execution is needed
to measure them.

\textbf{\SSv.} The screenshot audit covers 11 models: the three grounders above, Qwen2.5-VL-3B~\cite{bai2025qwen25vl}
and seven GRPO-trained 3B variants from \cref{sec:train}. A 40-unit pilot decided that the full audit was worthwhile,
and the 360-unit main sample gives the estimates in \cref{tab:ssv2}. Two independent reviewers agree closely with
each other and with our labels (\cref{tab:checks}). Where both reviewers disagreed with us, on 2 of 60 units, we
adopted their label. Outcome-correct clicks make up 6.4\% of the
rejected clicks. On web screenshots the share is 14.4\%, and mobile and desktop screenshots show about 1\%.
\Cref{fig:ssv2} shows the typical cases: a project logo and title that form one link, an address that is a single
link around the annotated word, a checkbox label, and a header link that duplicates a page button. The audit also
surfaced five items with a likely annotation problem, which we release with the audit.

\begin{table}[t]
  \centering
  \small
  \setlength{\tabcolsep}{4pt}
  \begin{tabular}{@{}lrrrrrc@{}}
    \toprule
    Split & Units & E-same & E-dup & N & U & E share [95\% CI] \\
    \midrule
    Web     & 145 & 8 & 12 & 119 & 6 & 14.4 [9.5, 21.2] \\
    Mobile  & 104 & 1 & 0  & 97  & 6 & 1.0 [0.2, 5.6] \\
    Desktop & 111 & 0 & 1  & 105 & 5 & 0.9 [0.2, 5.2] \\
    \midrule
    All     & 360 & 9 & 13 & 321 & 17 & 6.4 [4.3, 9.5] \\
    \bottomrule
  \end{tabular}
  \caption{\textbf{Screenshot audit of \SSv.} Box-rejected clicks of 11 models, judged blind. E share is
  (E-same + E-dup)/(E + N) in \% with Wilson intervals. Counting all U units as N or as E bounds the overall share
  between 6.1\% and 10.8\%.}
  \label{tab:ssv2}
\end{table}

\subsection{Stronger models, more misread errors}
\label{sec:strength}
\begin{figure}[t]
  \centering
  \includegraphics[width=\columnwidth]{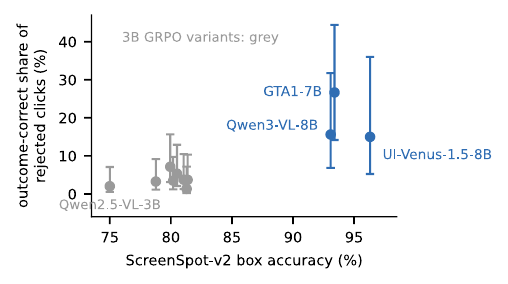}
  \caption{\textbf{The strongest grounders have the largest share of outcome-correct errors on \SSv.} Share of each
  model's audited box-rejected clicks judged equivalent, with Wilson intervals, against its box accuracy.}
  \label{fig:eshare}
\end{figure}
The share of outcome-correct clicks among rejections is not a constant of the benchmark. Among the 11 audited models
it is highest for the strongest grounders (\cref{fig:eshare}). Pooled over the audit units that contain a click of one of the three 7--8B grounders,
20.6\% [12.7, 31.6] of rejected clicks are outcome-correct. For units with clicks of the 3B models only, the share is
2.9\% [1.5, 5.6]. What remains for the strongest models is more often the other copy of the target, a label, or the
rest of a button. On the live applications the rate is high for every model, 3B models included, because duplicated
elements make up a large part of the items there. \Cref{sec:noise}
measures what this does to close comparisons.

\section{Where the Fix Belongs}
\label{sec:train}

\begin{figure*}[t]
  \centering
  \includegraphics[width=\textwidth]{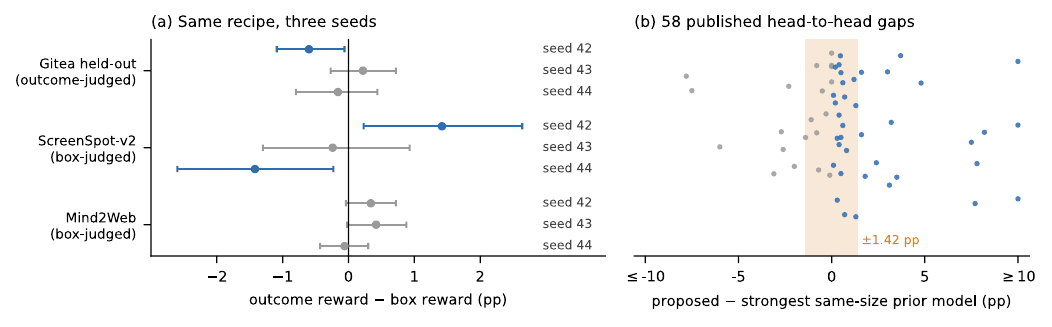}
  \caption{\textbf{Seeds and sampling set the resolution of \SSv.} (a) Outcome-reward minus box-reward accuracy for the
  same recipe trained with three seeds. Blue marks intervals that exclude zero (Gitea resamples pages, \SSv resamples
  screenshots, \MW resamples tasks). On \SSv seed 42 shows a significant gain and seed 44 an equally significant loss.
  (b) Head-to-head gaps between a proposed model and the strongest prior model of the same size, as reported on \SSv
  by 28 recent papers (blue: proposed model ahead). The band marks $\pm$1.42 points, the size of the single-seed effects in (a) that reverse across seeds.}
  \label{fig:noise}
\end{figure*}

Outcome maps also give a reward that credits the equivalent clicks they find. GUI-Libra~\cite{yang2026guilibra} argues that box
rewards are only partially verifiable and that the missing credit hurts training. If that held at our scale, the
correction would belong in the reward as well as in evaluation. GUI-Libra studies multi-step navigation; we test
whether the same ambiguity costs anything when training single-step grounding.

\textbf{Setup.} We split the Gitea items by page layout, giving 1,745 training and 1,841 held-out items from layouts
never seen in training. Items that the base model answers inside the box in all 8 samples carry no GRPO gradient under
any reward and are removed for all arms. This leaves 1,589 training items, and 313 of them have at least one
outcome-equivalent element. We fine-tune Qwen2.5-VL-3B-Instruct~\cite{bai2025qwen25vl} with LoRA~\cite{hu2022lora}
using the GRPO implementation of GTA1~\cite{yang2025gta1}, with 8 samples per item for 2 epochs (396 updates). Three
arms differ only in the reward. \emph{Box} rewards clicks inside the target box. \emph{Outcome} rewards clicks inside
the box of the target or of any outcome-equivalent element found by the outcome maps. This stand-in for executing
every rollout click agrees with the execution judge on 98.6\% of 14,727 held-out clicks of the eight 3B models, and
only 5 outcome-correct clicks fall outside every rewarded box. \emph{GUI-Libra} uses the box reward with KL
regularisation ($\beta = 0.001$) and success-adaptive scaling of negative advantages ($\lambda_0 = \kappa = 0.5$). All
arms share data, order and hyperparameters. We train three seeds for every arm, and
evaluate with greedy decoding by execution on the held-out Gitea layouts and on the Oscar shop, which no model saw in
training, and by box on \SSv and on the 4,731 \MW steps with element-level instructions.

\begin{table}[t]
  \centering
  \small
  \setlength{\tabcolsep}{3pt}
  \setlength{\tabcolsep}{2pt}
  \resizebox{\columnwidth}{!}{%
  \begin{tabular}{@{}lcccc@{}}
    \toprule
    & Gitea held-out & Oscar & \SSv & \MW \\
    & outcome & outcome & box & box \\
    \midrule
    Base model     & 78.4 & 91.0 & 75.0 & 83.7 \\
    Box reward     & 85.2 \tiny{\!\![85.2, 85.3]\!} & 89.6 \tiny{\!\![89.4, 89.9]\!} & 80.5 \tiny{\!\![80.0, 81.3]\!} & 86.5 \tiny{\!\![86.4, 86.6]\!} \\
    Outcome reward & 85.0 \tiny{\!\![84.6, 85.5]\!} & 89.7 \tiny{\!\![89.3, 90.2]\!} & 80.4 \tiny{\!\![78.8, 81.4]\!} & 86.7 \tiny{\!\![86.6, 86.9]\!} \\
    GUI-Libra      & 84.8 \tiny{\!\![84.5, 85.3]\!} & 90.0 \tiny{\!\![89.8, 90.2]\!} & 80.9 \tiny{\!\![80.5, 81.2]\!} & 86.7 \tiny{\!\![86.2, 87.0]\!} \\
    \midrule
    Outcome $-$ box & $-$0.18 & +0.06 & $-$0.08 & +0.23 \\
    \bottomrule
  \end{tabular}}
  \caption{\textbf{Reward definition and held-out accuracy (\%).} Each arm shows the mean over seeds 42, 43 and 44
  with the range across seeds, and the last row gives the difference of the outcome and box means in points.
  Per-seed differences with confidence intervals are in \cref{fig:noise}a.}
  \label{tab:train}
\end{table}

\textbf{Results.} Every arm improves on the base model by 6--7 points on held-out Gitea layouts, by 4--6 points on
\SSv and by about 3 points on \MW (\cref{tab:train}). On the unseen shop every arm ends 0.8 to 1.4 points below a
strong base model, so this training does not transfer there. The reward definition makes no stable difference.
Averaged over the three seeds, outcome and box rewards lie within a quarter of a point of each other on every test
set, and individual seeds point in both directions (\cref{fig:noise}a). Taking each training run as one observation,
the three seeds bound the difference to about one point on held-out Gitea and \MW. On \SSv the seed spread is too
large to bound it this tightly (\cref{sec:noise}). GUI-Libra lands in the same range, within half a point of the box
reward on average on every test set.

\textbf{Why the extra credit adds little.} The two rewards differ only on items with an equivalent element, about a
fifth of the training items. On the 96 held-out items whose instruction matches several elements with the same
outcome, the base model already reaches the target outcome on 92\%. Every trained model of every arm and seed reaches
100\%, while picking the annotated copy only 27--29\% of the time. The box reward on these items acts as label noise
that averages out. At this scale the richer reward has almost nothing left to improve, which is a ceiling of the test
rather than proof that such a reward can never help. On the shop, every model, the base model included, reaches the
target outcome on all 55 items with same-outcome copies.

\section{Bias and Noise at the One-Point Scale}
\label{sec:noise}

\begin{figure}[t]
  \centering
  \includegraphics[width=\columnwidth]{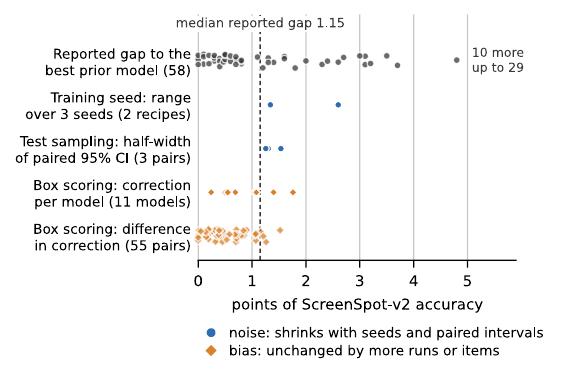}
  \caption{\textbf{Error budget of one-point comparisons on \SSv.} Each row shows one source of variation on the same
  scale as the margins that recent papers report. Seed ranges come from the three box-reward and the three
  outcome-reward runs of \cref{sec:train}, interval half-widths from the three pairs of 7--8B grounders, and
  corrections from the audit (point estimates; intervals in the supplement). The rows have different statistical
  meanings, so the figure compares scales and is not an additive decomposition.}
  \label{fig:budget}
\end{figure}

Three things move a close comparison on \SSv: the training seed, the test sample and the scoring rule
(\cref{fig:budget}). The first two are noise. More seeds average them out, and a paired interval bounds what remains.
The third is a bias, a one-sided under-count of correct clicks. It belongs to the model and the benchmark, so repeated runs and more items
from the same benchmark leave it in place.

\textbf{Seeds.} Retraining the same recipe with three seeds changes \SSv accuracy by up to 1.3 points for box rewards
and up to 2.6 points for outcome rewards. The same models vary by less than a point on \MW and on held-out Gitea
layouts. With seed 42 the outcome reward wins on \SSv with a
paired McNemar test~\cite{mcnemar1947} at $p = 0.03$, and with seed 44 the box reward wins at $p = 0.02$
(\cref{fig:noise}a). A single-seed significance test is a statement about one training run and says little about the
method.

\textbf{Test-set sampling.} Even a fixed model is measured on a finite sample of 1,272 items. We resampled \SSv items
10,000 times and computed paired intervals for the three strong grounders, the size class most papers compete in. The
half-width of the interval on a difference is 1.26--1.54 points. GTA1-7B and Qwen3-VL-8B differ by 0.31 points, and
their interval extends more than a point to each side of zero. The leads of UI-Venus-1.5-8B over both models, about
three points each, are clearly resolved.

\textbf{Scoring.}
\label{sec:rerank}
Outcome scoring raises \SSv accuracy by 0.2 to 1.8 points, depending on the model (\cref{tab:permodel} in the supplement). GTA1-7B gains
the most, so its gap to UI-Venus-1.5-8B shrinks from 2.9 to 1.7 points. When two models receive a similar correction,
their order stays and only both absolute scores move. What can change an order is the difference between two models'
corrections, which reaches 1.5 points for one pair. Three pairs change order in the point estimates, all within
0.8 points under box scoring, for example the seed-43 box-reward model and its outcome-reward twin (+0.24 under box
scoring, $-0.23$ corrected). These reversals are descriptive. Adding the audit uncertainty of the correction to the
box gap gives intervals that include zero for all three pairs (\cref{tab:pairs}), so the corrected order of these
pairs is not resolved either. Box scoring keeps the coarse leaderboard intact and decides the order of close models
partly by how often each model picks an equivalent element.

\textbf{Reported margins.} Two independent literature searches, each coded with fixed rules and checked against the
original tables, together cover 29 recent grounding papers that report \SSv, 28 of which contain a head-to-head
comparison. Of the 58 head-to-head comparisons with the strongest prior model of the same size, the median absolute
gap is 1.15 points (\cref{fig:noise}b). The proposed model leads in 38 of them, and in 21 of these the lead is no
larger than a single-seed effect. Ablations show a similar median of 1.2 points. Gains over the untrained base model
are larger, with a median of 4.0 points, so the field's progress is real and visible at that scale. No paper reports
run-to-run variation on \SSv.

All three sources act at the scale of the reported margins. Several seeds and a paired interval handle the noise.
Only a rule that credits equivalent clicks removes the bias.

\section{Recommendations and Outlook}
\label{sec:discussion}

\textbf{Recommendations.} Our measurements translate into two practices for evaluation that are cheap to adopt.
\begin{enumerate}
  \item \emph{Credit equivalent clicks on the web, and report both scores.} On web interfaces, between one in seven and
  one in three of the clicks that box scoring rejects do the right thing, and for the strongest grounders this is
  about a fifth of their remaining \SSv errors. Benchmarks can ship outcome maps or lists of equivalent elements with
  each web item. Reporting outcome-corrected
  accuracy next to box accuracy shows how much of a margin depends on the choice between equivalent targets.
  \item \emph{Report seeds and paired intervals.} Three seeds and a paired bootstrap interval cost little compared
  with training, and they turn a 1-point margin from an anecdote into evidence.
\end{enumerate}
In our 3B study, box labels remained an adequate training target (\cref{sec:train}).

\textbf{Where equivalence comes from.} The difference between web and native screenshots follows from how the two
are built. Web pages routinely expose one action through several elements: the image and title of a card, a label
bound to its control, a link repeated in a header and a page body. Native mobile and desktop widgets usually have a
single hit target, so a click outside the box is almost always a real error, and the audit finds about 1\% there
against 14.4\% on web screenshots. Lists of equivalent elements therefore matter most for web items.

\textbf{Scope.} Execution judging covers two live web applications, a code forge and an online shop that stand in for
the GitLab and shopping sites of WebArena, and the screenshot audit extends the picture to public mobile, desktop and
web data. The
training study uses a 3B model with LoRA and about 400 GRPO updates, and larger models or longer schedules may use the
extra credit differently. Equivalence is defined for one interface state, which matches how grounding benchmarks are
built. The released probe is designed to
extend to more applications, mobile platforms and multi-step tasks.

\textbf{Conclusion.} A click is right when it does the right thing. Graded this way, many of the errors that still
separate strong grounders on the web are correct clicks. This under-count does not average out, and it is as large as
the gaps the field now reports. Outcome maps and execution judging remove it from evaluation.

\clearpage
{
    \small
    \bibliographystyle{ieeenat_fullname}
    \bibliography{main}
}

\maketitlesupplementary
\label{supp:start}
\setcounter{section}{0}\renewcommand{\thesection}{\Alph{section}}\renewcommand{\theHsection}{supp.\Alph{section}}
\setcounter{table}{0}\renewcommand{\thetable}{S\arabic{table}}\renewcommand{\theHtable}{supp.S\arabic{table}}
\setcounter{figure}{0}\renewcommand{\thefigure}{S\arabic{figure}}\renewcommand{\theHfigure}{supp.S\arabic{figure}}

\noindent Sections A--I, Figure S1 and Tables S1--S7 are in this supplement. All other section, table and figure numbers refer to
the main paper.

\section{Probe and Outcome Maps}
\label{supp:probe}

\begin{figure*}[t]
  \centering
  \includegraphics[width=\textwidth]{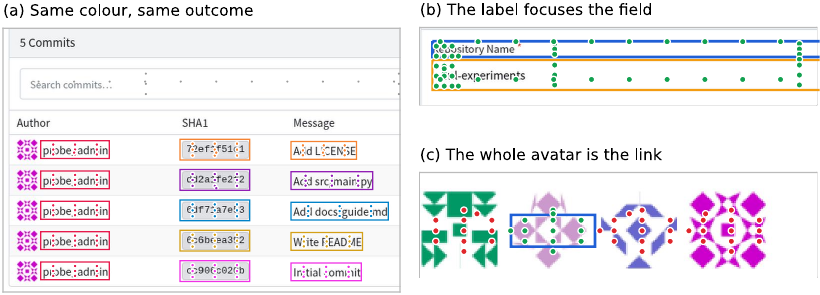}
  \caption{\textbf{Outcome maps of a self-hosted Gitea instance,} made by executing
  about a thousand clicks per page. (a) Clicks that produce the same outcome share a colour. Each commit is opened by
  its hash and by its message, and the author name in every row opens the same profile.
  (b) Clicking the label of a text field (blue box) focuses the field (orange). (c) The annotated box covers the
  middle of an avatar, and clicks anywhere on the avatar open the same profile. Green dots reach the outcome of the
  annotated target and red dots reach a different one. Box scoring counts every green dot outside the blue box as an
  error.}
  \label{fig:outmap}
\end{figure*}

\textbf{Signatures.} \Cref{tab:sig} lists the outcome signatures in order of precedence. A navigation records the
method, the URL with session tokens such as \texttt{\_csrf} removed, and the names of submitted form fields. A page
change is a difference between the post-hover and post-click snapshots in the fingerprints of visible nodes, the URL
after \texttt{pushState}, the focused text control, or the number of open dialogs.

\begin{table}[h]
  \centering
  \small
  \begin{tabular}{@{}cll@{}}
    \toprule
    Order & Signature & Recorded as \\
    \midrule
    1 & \texttt{nav} & main-frame navigation, aborted \\
    2 & \texttt{popup} & new window or tab \\
    3 & \texttt{req} & non-GET request, aborted \\
    4 & \texttt{dialog} & alert or confirm, dismissed \\
    5 & \texttt{dom} & visible change of the page \\
    6 & \texttt{noop} & none of the above \\
    \bottomrule
  \end{tabular}
  \caption{Outcome signatures.}
  \label{tab:sig}
\end{table}

\textbf{Waiting rule.} After the click the probe waits at least 250\,ms and then until requests issued after the click
have finished, for at most 1.5\,s more. Event streams, web sockets and URLs containing \texttt{/events},
\texttt{/stream}, \texttt{/sse}, \texttt{longpoll} or \texttt{/socket} are ignored. The first version of the probe also
waited for long-lived connections opened before the click. Gitea keeps such a connection open, so almost every click
waited the full 1.5\,s and probing took 3.4\,s per point. We changed only this rule, for speed. A preliminary
disagreement rate of about 26\% on the first 16 pages had been seen at that point, so the second version was accepted
only if it reproduced the outcomes of the first, independent of the direction of any result. The comparison covers 58
pages that both versions finished with at least 95\% repeatability. On these pages 99.4\% of 60,995 matched points
received the same outcome, and the overall disagreement rate changed by 0.1 points.

\textbf{Self-check.} Thirty disagreeing targets were drawn at random (seed 20261024) from the 85 pages with at least
95\% repeatability. Each was judged against written criteria from its signatures and equivalent elements, with
rendered overlays where needed. Twenty-eight are true disagreements, including panels (b) and (c) of
\cref{fig:outmap}. Two are false positives. In one, the Delete links of three different labels open one shared
confirmation dialog, and the page-change fingerprint merges the three actions. In the other, a search field already
holds focus after some page loads, so a click on it changes nothing. \Cref{tab:types} breaks down all 774 disagreeing
targets on these pages.

\textbf{Page statistics.} Of 5,073 interactive targets on 94 pages, 103 are dead. Among the 4,970 live targets, 748 have
a distinct equivalent, 127 have at least 10\% inert box area, and 881 (17.7\%) disagree with their box in at least one
way. Repeatability per page has a mean of 97.9\% and a minimum of 80.9\%. Nine pages fall below 95\%. Eight of them
are issue pages that embed Gitea's comment editor, and the ninth is the user list in site administration.

\textbf{Oscar.} The shop runs Django Oscar with SQLite in a separate environment. Fixed lists and templates populate
32 products (16 books and 16 home goods) across six categories, 48 programmatically generated placeholder images, 18
approved reviews, five accounts (one administrator and four customers) and seven pending orders. The seed also
includes a three-item basket, two saved addresses, a three-item wishlist, a stock alert, a site-wide 10\% discount
with a linked voucher, three shipping price bands, account messages and an About page. Product images are
$640\times480$ PNGs with coloured backgrounds and product-title text. A middleware placed first rolls back the database
transaction of every request, so that page views and analytics writes leave the database unchanged. Every page loads
with a 1.5\,s wait before probing. For this application the probe additionally normalises the boundary string of
multipart form submissions and the random element ids generated by the rich-text editor, and it treats a failed page
load as an error instead of a page state. The valid run followed two deployments that stopped on these signature
issues and on SQLite lock contention. Of 60 candidate pages, 56 completed with repeatability of at least 95\% (mean
99.8\%); four timed out during a later reload and count as failures. Execution judging covers 1,149 items with 11
models; three page-load errors leave 1,147 items for the core comparison.

\section{Execution Judging}
\label{supp:exec}

Items come from live targets with visible text, at most 40 per page. Of the 5,073 targets, 103 are dead and 443 have no
visible text, and the cap leaves 3,586 items on 94 pages. The role word in the instruction is chosen from the element type
(link, button, tab, checkbox, field). We stratify items by how many elements on the page match the instruction text.

\begin{table}[h]
  \centering
  \footnotesize
  \setlength{\tabcolsep}{3pt}
  \begin{tabular}{@{}lccc@{}}
    \toprule
    & Same-outcome & Different-outcome & Unique \\
    & copies (193) & copies (301) & (3,092) \\
    \midrule
    UI-Venus-1.5-8B & 35.2 / 95.9 & 32.9 / 32.2 & 90.0 / 88.6 \\
    GTA1-7B         & 34.2 / 100  & 35.2 / 33.9 & 88.5 / 87.0 \\
    Qwen3-VL-8B     & 28.5 / 90.7 & 29.6 / 28.9 & 87.8 / 86.5 \\
    Box centre      & 100 / 100   & 100 / 98.0  & 100 / 98.0 \\
    \bottomrule
  \end{tabular}
  \caption{Box / outcome accuracy (\%) by item type on Gitea.}
  \label{tab:strata}
\end{table}

\begin{table}[h]
  \centering
  \footnotesize
  \setlength{\tabcolsep}{2.5pt}
  \begin{tabular}{@{}llcccc@{}}
    \toprule
    & Model & Box & Outcome & $\Delta$ [95\% CI] & Rescued \\
    \midrule
    \multirow{3}{*}{Gitea} & UI-Venus-1.5-8B & 71.9 & 75.0 & +3.1 [1.7, 4.5] & 16.9\% \\
    & GTA1-7B & 69.4 & 72.6 & +3.2 [2.0, 4.5] & 15.2\% \\
    & Qwen3-VL-8B & 64.9 & 67.4 & +2.5 [1.1, 3.8] & 11.5\% \\
    \midrule
    \multirow{3}{*}{Oscar} & UI-Venus-1.5-8B & 79.6 & 85.9 & +6.3 [4.3, 8.3] & 35.9\% \\
    & GTA1-7B & 78.5 & 84.7 & +6.2 [4.5, 7.8] & 34.0\% \\
    & Qwen3-VL-8B & 73.9 & 82.3 & +8.4 [6.7, 10.3] & 34.7\% \\
    \bottomrule
  \end{tabular}
  \caption{Execution judging with natural instructions (accuracy in \%). ``Rescued'' is the share of box-rejected
  clicks that reach the target outcome. All 3,586 Gitea and 1,149 Oscar items were rewritten without exclusions.
  Pooled over models, the rescued share on items whose template text was unique is 6.8\% on Gitea (2.8\% with the
  template) and 30.6\% on Oscar (16.4\%).}
  \label{tab:nl}
\end{table}

Repeated execution of 712 predictions from a fresh page load agreed in 98.6\% of cases. Predictions that cannot be
parsed count as wrong under both scores (2.3\% of UI-Venus-1.5-8B outputs, none for the other two models).

\section{Screenshot Audit of \SSv}
\label{supp:audit}

\textbf{Units and sampling.} Clicks are mapped to page coordinates with the frozen evaluation convention (prediction
plus crop offset, box inclusive). Unparsed predictions are excluded. Units are sorted by item and position and shuffled
with seed 20261025. The pilot rule stated that the audit would stop with the conclusion ``\SSv is barely affected'' if
the pilot share fell below 5\%. The pilot gave 5.4\% (2 E, 35 N, 3 U), and the main sample followed.

\textbf{Review.} The 60 review units were drawn with seed 20261026. The reviewers worked independently from a
self-contained page that shows only the panel, the instruction and the category definitions. \Cref{tab:kappa} gives
the agreement over the three classes E, N and U. The rule to adopt the reviewers' label when both disagree with ours
was added after seeing the review. It changed two units, one from U to E-dup and one from N to U, and moved the
overall share from 6.1\% to 6.4\%.

\begin{table}[h]
  \centering
  \small
  \begin{tabular}{@{}lcc@{}}
    \toprule
    Pair & Agreement & Cohen's $\kappa$ \\
    \midrule
    Reviewer A vs reviewer B & 98.3\% & 0.95 \\
    Ours vs reviewer A & 96.7\% & 0.89 \\
    Ours vs reviewer B & 95.0\% & 0.84 \\
    \bottomrule
  \end{tabular}
  \caption{Agreement on the 60 review units.}
  \label{tab:kappa}
\end{table}

\textbf{Per-model results.} \Cref{tab:permodel} gives the share of equivalent clicks among each model's audited units
and the corrected accuracy, $\text{acc}_\text{box} + \text{err}_\text{box} \times \text{E share}$, with intervals from
resampling units.

\begin{table}[h]
  \centering
  \footnotesize
  \setlength{\tabcolsep}{1.8pt}
  \begin{tabular}{@{}lccccc@{}}
    \toprule
    Model & Units & E/N/U & Box & Corrected [95\% CI] & Rank \\
    \midrule
    UI-Venus-1.5-8B & 20 & 3/17/0 & 96.31 & 96.86 [96.31, 97.54] & 1 $\rightarrow$ 1 \\
    GTA1-7B & 33 & 8/22/3 & 93.40 & 95.16 [94.13, 96.29] & 2 $\rightarrow$ 2 \\
    Qwen3-VL-8B & 35 & 5/27/3 & 93.08 & 94.16 [93.34, 95.12] & 3 $\rightarrow$ 3 \\
    3B, outcome s42 & 84 & 3/78/3 & 81.37 & 82.06 [81.37, 82.93] & 4 $\rightarrow$ 4 \\
    3B, box s43 & 78 & 1/74/3 & 81.29 & 81.53 [81.29, 82.08] & 5 $\rightarrow$ \textbf{7} \\
    3B, outcome s43 & 82 & 3/77/2 & 81.05 & 81.76 [81.05, 82.64] & 6 $\rightarrow$ \textbf{5} \\
    3B, GUI-Libra s42 & 80 & 4/71/5 & 80.50 & 81.54 [80.74, 82.63] & 7 $\rightarrow$ \textbf{6} \\
    3B, box s44 & 89 & 3/83/3 & 80.19 & 80.87 [80.19, 81.73] & 8 $\rightarrow$ \textbf{9} \\
    3B, box s42 & 73 & 5/65/3 & 79.95 & 81.35 [80.25, 82.63] & 9 $\rightarrow$ \textbf{8} \\
    3B, outcome s44 & 96 & 3/89/4 & 78.77 & 79.46 [78.77, 80.33] & 10 $\rightarrow$ 10 \\
    Qwen2.5-VL-3B & 103 & 2/97/4 & 75.00 & 75.50 [75.00, 76.30] & 11 $\rightarrow$ 11 \\
    \bottomrule
  \end{tabular}
  \caption{\SSv audit per model (accuracy in \%). The correction only credits rejected clicks, since accepted clicks with a different outcome cannot be seen on screenshots. Ranks in bold change.}
  \label{tab:permodel}
\end{table}

\textbf{Pairs whose gap changes.} \Cref{tab:pairs} lists the three pairs that change order together with the
comparisons discussed in the main text. Eight of the 55 intervals on the difference in correction exclude zero, where
about three would be expected by chance. The seed-42 comparison of outcome and box rewards, significant under box
scoring, halves under outcome scoring.

\begin{table}[h]
  \centering
  \scriptsize
  \setlength{\tabcolsep}{1.5pt}
  \begin{tabular}{@{}lccc@{}}
    \toprule
    Pair (A vs B) & Box & $\Delta$ corr.\ [95\% CI] & Corrected [95\% CI] \\
    \midrule
    box s44 vs box s42 & +0.24 & $-$0.71 [$-$1.60, $-$0.07] & $-$0.48 [$-$1.36, 0.17] \\
    box s43 vs outcome s43 & +0.24 & $-$0.47 [$-$1.46, 0.41] & $-$0.23 [$-$1.22, 0.65] \\
    box s43 vs GUI-Libra & +0.79 & $-$0.80 [$-$1.77, $-$0.01] & $-$0.01 [$-$0.98, 0.78] \\
    outcome s42 vs box s42 & +1.42 & $-$0.71 [$-$1.59, $-$0.06] & +0.70 [$-$0.17, 1.36] \\
    box s43 vs box s42 & +1.34 & $-$1.16 [$-$2.34, $-$0.26] & +0.18 [$-$1.00, 1.08] \\
    UI-Venus vs GTA1 & +2.91 & $-$1.21 [$-$2.45, $-$0.01] & +1.70 [0.46, 2.90] \\
    UI-Venus vs Qwen3-VL & +3.22 & $-$0.53 [$-$1.62, 0.48] & +2.70 [1.60, 3.70] \\
    GTA1 vs Qwen3-VL & +0.31 & +0.68 [$-$0.50, 1.90] & +0.99 [$-$0.19, 2.21] \\
    \bottomrule
  \end{tabular}
  \caption{Selected pairs on \SSv (points). The corrected-gap interval adds the box gap to the interval of the difference
  in correction, so it reflects audit uncertainty and holds the test set fixed. All 55 pairs are released.}
  \label{tab:pairs}
\end{table}

\textbf{Likely annotation problems.} Seven units on five items were flagged during judging. On three items the click
follows the instruction and the annotated box marks another element, for example ``switch the temperature display to
Fahrenheit'' boxed on \emph{Notifications}. On the other two the instruction fits both the click and the box. All seven
units count as N in every estimate and are listed in the release.

\section{\MW Static Analysis}
\label{supp:m2w}

For each click step we evaluate the $1280\times720$ window that the grounders see and look for elements that share the
target's outcome from page annotations alone: elements with the same link target, and labels bound to the target
control. The released page HTML keeps no link targets, so only the label channel can be measured, and the reported
1.3\% is a lower bound. Step-level instructions name the element and its text, and task-level instructions give the
task and previous actions.

\section{Training Details}
\label{supp:train}

\textbf{Recipe.} GTA1 GRPO (commit \texttt{81f0c4c}) with gradient checkpointing, a shared coordinate parser for reward
and evaluation, and a chunked vision attention patch to fit 24\,GB GPUs. LoRA has rank 16 and $\alpha = 32$ on the
attention and MLP projections, with the vision modules frozen. Learning rate $2\times10^{-5}$ with 5\% warm-up and linear decay, 8 items per
update on 8 GPUs, 8 samples per item at temperature 1, at most 64 new tokens, images up to $1280\times720$. Each arm
takes about one hour on eight RTX 4090 GPUs. GUI-Libra does not give values for $\lambda_0$ and $\kappa$, and we use
0.5 for both.

\textbf{Decision rules.} The first study (seed 42) asked whether the outcome reward raises outcome accuracy on held-out
Gitea layouts, with a guard against losing more than one point on \SSv. The difference was $-0.60$ points
[$-1.09$, $-0.06$], so the outcome reward gave no gain. Seed 42 did show a $+1.42$ point difference on \SSv, which
motivated a confirmation with seeds 43 and 44 on \SSv and \MW. Averaged over the two new seeds the difference is
$-0.83$ [$-1.66$, $+0.04$] on \SSv. On \MW it is $+0.18$ [$-0.11$, $+0.47$]. The seed-42 effect therefore did not
replicate. Seeds 43 and 44 were also evaluated on held-out Gitea as a report-only analysis, which gives the three-seed
Gitea results in \cref{tab:train}. GUI-Libra seeds 43 and 44 were added later as a report-only analysis, with the same
script, data and hyperparameters. Averaged over three seeds, GUI-Libra minus box is $-0.36$ (Gitea), $+0.40$ (Oscar),
$+0.47$ (\SSv) and $+0.16$ (\MW) points.

\section{Test-Set Noise}
\label{supp:testnoise}

We recomputed per-item correctness for all 11 models from the raw predictions and checked it against the audit totals.
Pairs are resampled jointly over items (10,000 times, seed 20261025).

\begin{table}[h]
  \centering
  \footnotesize
  \setlength{\tabcolsep}{2.5pt}
  \begin{tabular}{@{}lcccc@{}}
    \toprule
    Pair & Diff & 95\% CI & Discord. & McNemar $p$ \\
    \midrule
    GTA1 $-$ Venus & $-$2.91 & [$-$4.25, $-$1.65] & 5.7\% & $<$0.001 \\
    Qwen3 $-$ Venus & $-$3.22 & [$-$4.48, $-$1.97] & 5.3\% & $<$0.001 \\
    GTA1 $-$ Qwen3 & +0.31 & [$-$1.26, 1.81] & 8.0\% & 0.77 \\
    \bottomrule
  \end{tabular}
  \caption{Paired comparisons of 7--8B grounders on \SSv (points). Discord.\ is the share of items that exactly one
  of the two models gets right.}
  \label{tab:testnoise}
\end{table}

\section{Literature Coding}
\label{supp:lit}

We collected papers that report \SSv grounding accuracy and coded three kinds of comparison: the proposed model
against the strongest prior model of the same size reported in the same table (head-to-head), against its own
untrained base model, and ablations. Two searches were run independently with their own coding notes, and every
number was checked against the original table. The union contains 29 papers, 58 head-to-head comparisons from 28 of
them, 47 base comparisons and 166 ablation rows. The reference thresholds are the single-seed effect of $1.42$ points
and the largest seed spread of $2.60$ points from \cref{sec:noise}. Of the head-to-head gaps, 57\% are at most 1.42 and
69\% at most 2.60 points. The paper list and coded rows are released.

\section{Protocol Log}
\label{supp:log}

Each analysis has a rule file with a timestamp and checksum. Later changes were appended as amendments that state
what had been seen when they were written. The amendments behind the reported analyses are, in order:
\begin{enumerate}
  \item The probe waiting rule of \cref{supp:probe}.
  \item Execution-judged items taken from the second probe run, decided before any judging.
  \item Qwen3-VL-8B added after its adapter reproduced the published \SSv accuracy (93.1\% against 91.7\%).
  \item The coordinate parser extended to parenthesised boxes, written when only base-model outputs existed.
  \item The report-only Gitea evaluation of seeds 43 and 44.
  \item The audit pilot outcome and the exploratory flags.
  \item The reviewer adjudication rule.
  \item The pairwise correction analysis of \cref{sec:rerank}.
  \item The natural-instruction check of \cref{sec:eval} and its reading rule, written before any rewrite. The
  rewriting model was then changed from Qwen2.5-7B-Instruct, which would have had to be downloaded, to
  Qwen3-4B-Instruct-2507, which was available locally, still before any rewrite.
  \item GUI-Libra seeds 43 and 44, with a fallback to four GPUs at the same 64 samples per update. Both seeds ran on
  eight GPUs.
\end{enumerate}
The first Gitea execution of the natural-instruction check failed the drift rule (52\% of box centres changed),
because the login session of the probe had expired. The run was declared invalid as the rule requires, the session was
renewed, and the execution was repeated with drift 1.8\%. Earlier Gitea executions passed the same check.
Item 4 also resolved the low parse rate that stopped the first training diagnostic. A memory overflow in the same
diagnostic was resolved by the vision attention patch, which matches the upstream attention to numerical precision.

\section{Release}
\label{supp:release}

We release the probe and its tests, the seeded Gitea application, all outcome maps, the 3,586 execution-judged items
with every prediction and outcome, the \SSv audit panels with our labels and both reviewers' labels, training and
evaluation code with all checkpoints' predictions, the literature coding, and every rule file with its checksum.

\end{document}